\documentclass[letterpaper]{article} 
\usepackage{aaai2026}  
\usepackage{times}  
\usepackage{helvet}  
\usepackage{courier}  
\usepackage[hyphens]{url}  
\usepackage{graphicx} 
\usepackage{natbib}  
\usepackage{caption} 
\usepackage{algorithm}
\usepackage{algorithmic}
\usepackage{amsmath}
\usepackage{amssymb}
\usepackage{booktabs}
\usepackage{multirow}
\usepackage{tcolorbox}
\usepackage{colortbl}
\usepackage{tabularx}
\definecolor{lightgray}{gray}{0.92}

\usepackage{newfloat}
\usepackage{listings}
\usepackage{xcolor}
\DeclareCaptionStyle{ruled}{labelfont=normalfont,labelsep=colon,strut=off} 
\floatstyle{ruled}
\newfloat{listing}{tb}{lst}{}
\floatname{listing}{Listing}
\title{\textsc{S2D-Align}: Shallow-to-Deep Auxiliary Learning for\\Anatomically-Grounded Radiology Report Generation}
\author{
    Jiechao Gao\textsuperscript{\rm 1,}\thanks{Corresponding author}, Chang Liu\textsuperscript{\rm 2}, Yuangang Li\textsuperscript{\rm 3}
}
\affiliations{
    \textsuperscript{\rm 1}Stanford University\\
    \textsuperscript{\rm 2}University of Science and Technology of China\\
    \textsuperscript{\rm 3}University of California, Irvine\\

    jiechao@stanford.edu, christzhaung@gmail.com, yuanganl@uci.edu
    
}

\usepackage{bibentry}

\begin{document}

\maketitle

\begin{abstract}
Radiology Report Generation (RRG) aims to automatically generate diagnostic reports from radiology images.
To achieve this, existing methods have leveraged the powerful cross-modal generation capabilities of Multimodal Large Language Models (MLLMs), primarily focusing on optimizing cross-modal alignment between radiographs and reports through Supervised Fine-Tuning (SFT).
However, by only performing instance-level alignment with the image-text pairs, the standard SFT paradigm fails to establish anatomically-grounded alignment, where the templated nature of reports often leads to sub-optimal generation quality.
To address this, we propose \textsc{S2D-Align}, a novel SFT paradigm that establishes anatomically-grounded alignment by leveraging auxiliary signals of varying granularities.
\textsc{S2D-Align} implements a shallow-to-deep strategy, progressively enriching the alignment process: it begins with the coarse radiograph-report pairing, then introduces reference reports for instance-level guidance, and ultimately utilizes key phrases to ground the generation in specific anatomical details.
To bridge the different alignment stages, we introduce a memory-based adapter that empowers feature sharing, thereby integrating coarse and fine-grained guidance.
For evaluation, we conduct experiments on the public \textsc{MIMIC-CXR} and \textsc{IU X-Ray} benchmarks, where \textsc{S2D-Align} achieves state-of-the-art performance compared to existing methods.
Ablation studies validate the effectiveness of our multi-stage, auxiliary-guided approach, highlighting a promising direction for enhancing grounding capabilities in complex, multi-modal generation tasks.
\end{abstract}


\section{Introduction}

Medical imaging, such as X-rays and Computed Tomography (CT), serves as an indispensable non-invasive tool in modern diagnostics, offering a crucial way to visualize the internal structures of human body conditions.
Following the interpretation of these images, radiologists are required to record detailed diagnostic reports that translate complex visual findings into precise medical language, forming a critical basis for subsequent clinical decision-making.
This manual process, however, is not only time-consuming but also susceptible to errors and omissions, particularly for less experienced radiologists, which can potentially degrade the quality of patient care.
To mitigate these challenges, the task of Radiology Report Generation (RRG) has been motivated by recent studies \cite{jing-etal-2018-automatic,nips-2018-hrgr-agent,chen-etal-2020-generating, liu2021exploring,chen-etal-2021-cross-modal,qin-song-2022-reinforced}, aiming to develop automatic solutions to alleviate the workload of radiologists, where this research direction has raised great attention from the communities of both artificial intelligence and clinical medicine.

Recent breakthroughs in Large Language Models (LLMs) \cite{touvron2023llama} have motivated Multimodal Large Language Models (MLLMs) \cite{zhu2023minigpt4,liu-etal-2023-llava} as the cornerstone for RRG, effectively overcoming the alignment challenges inherent in earlier methods \cite{liu-etal-2023-llava} trained from scratch on limited datasets.
Adapting these general MLLMs for the medical domain primarily involves two competing strategies, i.e., In-Context Learning (ICL) and Supervised Fine-Tuning (SFT).
ICL methods \cite{yan2023style}, which keep the LLM parameters frozen, typically rely on external annotators like RadGraph \cite{jain2021radgraph} to convert visual information into structured text (e.g., entities and relations), upon which few-shot demonstrations guide the generation.
However, their performance is highly sensitive to the quality of these text-based representations and the choice of demonstration examples, limiting their robustness in complex clinical scenarios.
Consequently, SFT has emerged as the dominant paradigm, establishing end-to-end alignment by directly fine-tuning the MLLM on radiograph-report pairs \cite{liu2024bootstrapping,wang-etal-2025-llm-rg4, hyland2023maira1,tu2023medpalm}.
Despite its prevalence, the standard SFT framework faces a critical bottleneck, where it performs alignment only at a coarse granularity between the entire image and its corresponding report.
This coarse-grained approach, confounded by the templated and often redundant nature of radiology reports, fails to establish precise correspondence between specific pathological findings and their anatomical locations.
This deficiency in alignment granularity directly undermines the factual correctness and clinical reliability of the generated reports.
Architecturally, this limitation is often exacerbated by the use of simple projection layers that bridge the visual encoder and the LLM, which are insufficient for learning fine-grained, region-to-text mappings for RRG.
Therefore, developing an effective method for \textit{anatomically-grounded alignment} has become a critical challenge for trustworthy RRG, where this pivotal problem motivates our work in this paper.

To address this critical challenge, we introduce \textsc{S2D-Align}, a novel fine-tuning paradigm designed to explicitly establish anatomically-grounded alignment.
At its core, we propose \textbf{Progressive Anatomical Grounding (PAG)}, a shallow-to-deep SFT strategy that systematically enriches the alignment process by leveraging auxiliary signals of varying granularities.
This multi-stage process begins with coarse-grained radiograph-report alignment, then incorporates reference reports for enhanced contextual understanding, and culminates in fine-grained grounding using clinically-relevant key phrases to connect text to specific anatomical regions.
To unify the learning signals across the diverse stages of PAG, we introduce the \textbf{Shallow-to-Deep Memory Adapter (SMA)}, a lightweight yet effective memory-based adapter that facilitates feature sharing and integrates coarse- and fine-grained guidance into a cohesive representation.
Our extensive experiments on the \textsc{IU X-Ray} and \textsc{MIMIC-CXR} benchmarks demonstrate that \textsc{S2D-Align} achieves new state-of-the-art performance compared to prevailing methods.
Generally speaking, the contributions of \textsc{S2D-Align} are threefold:
\begin{itemize}
\item We propose \textbf{Progressive Anatomical Grounding (PAG)}, an innovative multi-stage SFT framework that explicitly targets anatomically-grounded RRG;
\item We design the \textbf{Shallow-to-Deep Memory Adapter (SMA)}, an effective module that enable multi-grained feature sharing during the fine-tuning of MLLMs;
\item We conduct comprehensive experiments that not only validate the superiority of \textsc{S2D-Align}, but also highlight a promising direction for building more factually reliable and clinically trustworthy generative models.
\end{itemize}

\begin{figure*}[t!]
\centering
\includegraphics[width=\linewidth]{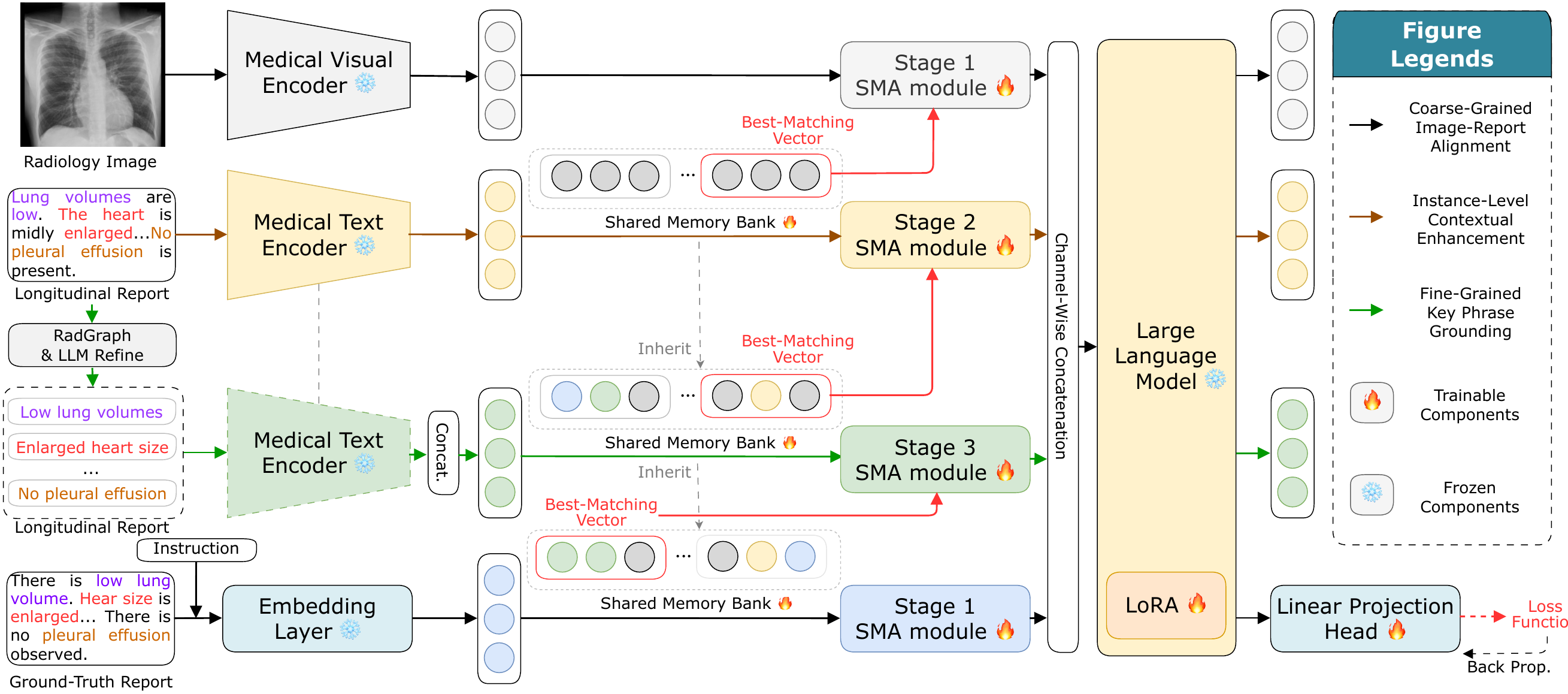}
\vspace{-2em}
\caption{\textbf{Overview of \textsc{S2D-Align}},
with the Progressive Anatomical Grounding (PAG) and Shallow-to-Deep Memory Adapter (SMA) modules as its core components.
Herein, we use the same medical text encoders to convert reference reports or key phrases into embeddings, and adopt a shared memory bank inherited from earlier stages to later ones throughout PAG.
}
\vspace{-1.8em}
\label{fig: pipeline}
\end{figure*}

\section{Related Work}

The advent of deep learning has catalyzed a significant paradigm shift in RRG over the last decade.
Foundational approaches \cite{jing-etal-2018-automatic,nips-2018-hrgr-agent,liu2021exploring,liu-etal-2021-contrastive,nicolson2022warmstart,huang2023kiut,tanida2023interactive,jin2024promptmrg} established the encoder-decoder framework, typically by training task-specific neural networks on benchmark datasets \cite{iu-xray,johnson2019mimic}.
These models primarily focused on enhancing cross-modal alignment to improve report quality, employing techniques such as memory networks \cite{chen-etal-2020-generating,chen-etal-2021-cross-modal}, attention mechanisms \cite{liu2021exploring}, reinforcement learning \cite{qin-song-2022-reinforced}, etc.
However, these methods, trained from scratch on limited-scale medical datasets, were fundamentally constrained in model capacity, limiting their applicability to complex, real-world clinical scenarios.
The emergence of Large Language Models (LLMs), pre-trained on massive text corpora, has introduced powerful text generation capabilities, motivating a new research direction to overcome the limitations of earlier methods.
Consequently, the dominant paradigm has shifted towards adapting these models for RRG by aligning a visual encoder with an LLM and performing Supervised Fine-Tuning (SFT) on radiograph-report pairs \cite{hyland2023maira1,tu2023medpalm,liu2024bootstrapping,wang-etal-2025-llm-rg4}.
Nevertheless, this standard SFT paradigm conducts at an instance-level of alignment, failing to establish the fine-grained mappings between specific visual findings and their textual descriptions necessary for anatomical grounding.
Among the most relevant works, LLM-RG4 \cite{wang-etal-2025-llm-rg4} attempts to address this by introducing an adaptive token fusion module and a token-level loss weighting strategy to prioritize descriptions of local regions.
Yet, its learning process is still fundamentally constrained by instance-level data pairs, lacking explicit anatomical guidance.
In contrast, our proposed \textsc{S2D-Align} directly tackles this challenge by injecting explicit, multi-grained anatomical signals---such as key phrases and their corresponding visual regions---into the SFT process to progressively achieve anatomically-grounded alignment.

\section{Methodology}

In this section, we detail the architecture and training methodology of \textsc{S2D-Align}.
As illustrated in Figure \ref{fig: pipeline}, our framework is built upon three core modules, i.e., a frozen medical visual encoder ($\mathcal{E}_v$), our proposed \textbf{Shallow-to-Deep Memory Adapter (SMA)}, and a Large Language Model (LLM) decoder ($\mathcal{G}_{LLM}$).
The fine-tuning of these modules is organized by our central contribution, the \textbf{Progressive Anatomical Grounding (PAG)} strategy, which leverages auxiliary signals to guide the model towards anatomically-grounded alignment.

The PAG strategy formalizes the fine-tuning as a three-stage curriculum.
At each stage $i$, the model is conditioned on a progressively enriched multi-modal context $C^{(i)}$. Let $I$ be the input radiograph, $R_{ref}$ be the reference report, and $K$ be the set of key phrases. We define the contexts as follows:
\begin{itemize}
    \item \textbf{Stage 1 (Coarse Alignment):} The context contains only the visual information distilled by the SMA.
    \begin{equation}
        C^{(1)} \triangleq \text{SMA}_v(\mathcal{E}_v(I))
    \end{equation}
    \item \textbf{Stage 2 (Contextual Enhancement):} The context is augmented with features from the reference report.
    \begin{equation}
        C^{(2)} \triangleq \texttt{concat}\left( C^{(1)}, \text{SMA}_{\text{t}}(\mathcal{E}_{\text{text}}(R_{ref})) \right)
    \end{equation}
    \item \textbf{Stage 3 (Fine-grained Grounding):} The context is further enriched with key phrase features.
    \begin{equation}
        C^{(3)} \triangleq \texttt{concat}\left( C^{(2)}, \text{SMA}_{\text{p}}(\mathcal{E}_{\text{text}}(K)) \right)
    \end{equation}
\end{itemize}
where $\text{SMA}_v$, $\text{SMA}_t$, and $\text{SMA}_p$ are the memory-based adapter modules for vision, reference reports, and key phrases, respectively, and $\texttt{concat}(\cdot)$ denotes the operation of concatenation along the channel dimension.
Given this formulation, the training objective for the $i$-th stage of PAG is to minimize the auto-regressive cross-entropy loss over the ground-truth report $R_{gt}$:
\begin{equation}
    \mathcal{L}^i_{\text{PAG}} = - \sum_{t=1}^{\left|R_{gt}\right|} \log p_{\Theta_i}\left(w_t | w_{<t}, C^{\left(i\right)}\right)
    \label{eq:pag_loss_stage_new}
\end{equation}
where $w_t$ is the $t$-th token of $R_{gt}$, and $p_{\Theta_i}$ is the probability predicted by the model with parameters $\Theta_i$ trainable at stage $i$. 
During inference, we discard the auxiliary signals ($C^{(i)}, i \geq 2$) and generate the report $R$ conditioned solely on the visual context $C^{(1)}$, which is formally expressed as:
\vspace{-0.5em}
\begin{equation}
    \vspace{-0.5em}
    R = \mathcal{G}_{\text{LLM}}\left( \cdot | C^{(1)} \right)
    \label{eq:overall_framework_new}
\end{equation}
This training-inference asymmetry is a key design principle in this work, enabling the model to learn from a multi-modal context while maintaining the efficiency of a standard RRG pipeline.
In the subsequent sections, we first introduce the visual feature extraction process, then illustrate the architecture of the SMA, and finally detail the PAG strategy.

\vspace{-0.2em}
\subsection{Visual Encoder}
\vspace{-0.2em}

The visual backbone of our framework is a pre-trained medical visual encoder, $\mathcal{E}_v$, which aims to encode an input radiology image $I \in \mathbb{R}^{H \times W \times C}$ into a sequence of feature embeddings.
We adopt the Vision Transformer (ViT) architecture \cite{dosovitskiy-etal-2021-vit}, which processes the image by partitioning it into a sequence of non-overlapping patches.
Latter, these patches are then embedded, incorporating positional information, to produce a sequence of patch-level feature vectors $V = \{v_1, v_2, \dots, v_N\}$, which are then adopted different stages throughout the entire process of PAG.

\vspace{-0.2em}
\subsection{Shallow-to-Deep Memory Adapter (SMA)}
\vspace{-0.2em}

As is noted above, a pivotal component in standard MLLMs is the connector module that bridges the visual encoder and the LLM.
Existing MLLMs typically employ simple connector designs like an MLP or with a further integration of Q-Former \cite{li2023blip2}, to project visual features into the embedding space of the LLM.
However, these approaches are insufficient for establishing the anatomically-grounded alignment in the context of RRG, due to two main reasons, where the insufficient capabilities of MLPs and failure of feature sharing across modalities to capture complex relationships and complementary information.
To address these shortcomings, we introduce the \textbf{Shallow-to-Deep Memory Adapter (SMA)}, a novel and efficient module designed to foster deep and interactive cross-modal alignment.
Unlike conventional connectors, the SMA operates based on a multi-head cross-attention mechanism, along with a \textit{memory bank} as a collection of $N_{\text{mem}}$ learnable query vectors, denoted as $Q_{\text{mem}} \in \mathbb{R}^{N_{\text{mem}} \times D_v}$.
During training, these memory queries dynamically interact with $V$ from the encoder, adaptively attending to and distilling the most salient visual information into a compact representation.
More importantly, the same memory bank is shared across all distinct alignment stages of PAG, enabling implicit feature sharing and compelling the adapter to learn a holistic and highly informative representation.
Given the feature $F_{aux}$ of the auxiliary signals (such as features of radiograph, reference report, and key phrases), this process is formally expressed as:
\vspace{-0.2em}
\begin{equation} \label{eq:sma_attn}
\vspace{-0.2em}
F_{\text{mem}} = \texttt{CrossAttn}(Q_{\text{mem}}, F_{aux}, F_{aux}; \Theta_{\text{SMA}})
\end{equation}
where $F_{\text{mem}} \in \mathbb{R}^{N_{mem} \times D_{aux}}$ is the resulting memory-enhanced feature that is fed to the LLM, and $D_{aux}$ varies according to different modalities.

\vspace{-0.2em}
\subsection{Progressive Anatomical Grounding (PAG)}
\vspace{-0.2em}

The core contribution of our approach is the Progressive Anatomical Grounding (PAG) strategy, which addresses the limitations of standard instance-level SFT.
PAG is conceptually motivated by Curriculum Learning (CL) \cite{Bengio2009CurriculumL}, a training paradigm that advocates for presenting easier examples to a model before progressively introducing more complex ones, thereby improving convergence and generalization.
However, a direct application of CL to RRG is challenging, as defining a meaningful difficulty metric for radiograph-report pairs is non-trivial, yet simple heuristics, such as report length or the number of findings, often fail to capture trustworthy clinical complexity and the required level of grounding.
To circumvent this challenge, PAG redefines the notion of difficulty not at the level of individual data samples, but the alignment task itself, which naturally fits the SFT nature of existing MLLMs.
In doing so, it performs a multi-stage curriculum that progressively increases the required alignment granularity, guiding the model from learning instance-level semantics to anatomically-grounded descriptions.
This is achieved by gradually introducing auxiliary textual signals of varying granularities across three individual stages, as detailed subsequently.

\paragraph{Stage 1: Coarse-Grained Image-Report Alignment.}
This initial stage of PAG establishes the basic alignment between the visual and textual modalities.
Given an input radiograph $I \in \mathbb{R}^{H \times W \times C}$ and its ground-truth report $R_{gt} = \{w_1, w_2, \dots, w_{|R_{gt}|}\}$, the forward pass proceeds as follows.
First, the frozen medical visual encoder $\mathcal{E}_v$ processes the image to extract a sequence of patch-level feature embeddings $V \in \mathbb{R}^{N \times D_v}$.
Next, these visual features $V$ are fed into our lightweight SMA, whose parameters $\Theta_{\text{SMA}}$ are the sole target to update in this stage.
The SMA comprises a series of learnable memory vectors, denoted as $Q_{\text{mem}} \in \mathbb{R}^{N_{\text{mem}} \times D_v}$, and levearges $Q_{\text{mem}}$ to adaptively distill the visual representation $V$ into a memory-enhanced one $V_{\text{mem}} \in \mathbb{R}^{N_{\text{mem}} \times D_v}$ via a cross-attention mechanism, written as:
\vspace{-0.2em}
\begin{equation} \label{eq:sma_cross_attn}
    \vspace{-0.2em}
    V_{\text{mem}} = \text{SMA}_v(Q_{\text{mem}}, V, V; \Theta_{\text{SMA}_v})
\end{equation}
$V_{\text{mem}}$ is then concatenated with the token embedding $E_{<t} = \text{Embed}(w_{<t})$ of $R_{gt}$ along the sequence dimension, with the resulting representation serving as the visual context for the LLM $\mathcal{G}_{\text{LLM}}$, which eventually predicts the probability distribution over the vocabulary for the next token $w_t$, conditioned on $V_{\text{mem}}$ and $w_{<t}$, written as:
\begin{equation}
    p(w_t | w_{<t}, V; \Theta_{\text{SMA}_v}) = \mathcal{G}_{\text{LLM}}\left(w_{<t}, V_{\text{mem}}\right)
\end{equation}
The training objective is to minimize the standard auto-regressive Cross-Entropy loss (CE), which maximizes the likelihood of the ground-truth report, written as:
\vspace{-0.2em}
\begin{equation} \label{eq:stage_1_loss}
    \vspace{-0.2em}
    \mathcal{L}^1_{\text{PAG}} = - \sum_{t=1}^{|R_{gt}|} \log p(w_t | w_{<t}, V; \Theta_{\text{SMA}_v})
\end{equation}
By optimizing this objective, we exclusively update the parameters $\Theta_{\text{SMA}_v}$, effectively aligning the feature space of the visual encoder with that of the LLM at an instance-level.

\paragraph{Stage 2: Instance-level Contextual Enhancement.}
To mitigate the ambiguity caused by the templated nature of radiology reports, this stage further introduces a reference report to enhance the instance-level context.
To implement this, we leverage the inherent longitudinal nature of the MIMIC-CXR dataset. For a given radiograph-report pair $(I, R_{\text{gt}})$ from a specific patient study, we select the reference report $R_{\text{ref}}$ from a different study of the same patient.
This strategy is clinically motivated, where serial radiographs of the same individual normally share a high degree of anatomical correspondence, yet their reports often differ based on subtle but diagnostically critical interval changes. 
Therefore in this stage, the overall pipeline is then tasked to generate the correct report $R_{gt}$ conditioned on both $I$ and $R_{\text{ref}}$, forcing it to discover case-specific visual cues.
To bridge $R_{\text{ref}}$ with the input of $I$, we adopts an BERT-based text encoder model to convert $R_{\text{ref}}$ into the corresponding representation $E_{\text{ref}} \in \mathbb{R}^{|R_{\text{ref}}| \times D_t}$, and utilizes a lightweight text adapter, architecturally identical to our SMA but with separate parameters ($\Theta_{\text{SMA}_t}$), to project the text embedding $E_{\text{ref}}$ into the shared feature space.
Crucially, while the parameters of the text adapter are distinct, the concatenated features are eventually processed in the context of the same memory queries $Q_{\text{mem}}$, ensuring consistent feature integration.
Then, the resulting representation is concatenated with $V_{\text{mem}}$ to form the input of the LLM $\mathcal{G}_{\text{LLM}}$, which eventually predicts the probability distribution similar to that of Eq. \ref{eq:stage_1_loss}, with the training objective $\mathcal{L}^2_{\text{PAG}}$ formulated by:
\vspace{-0.4em}
\begin{equation} \label{eq:stage_2_loss}
    \vspace{-0.5em}
    \mathcal{L}^2_{\text{PAG}} = - \sum_{t=1}^{|R_{gt}|} \log p(w_t | w_{<t}, V, R_{\text{ref}}; \Theta_{\text{SMA}_v}, \Theta_{\text{SMA}_t})
\end{equation}
Note that both SMA modules share the same memory vectors $Q_{\text{mem}}$ in this stage, where the additional reference report guides the model to develop a more robust instance-level understanding by contrasting against similar cases.

\begin{table*}[t]
    \centering
    \footnotesize
    \setlength{\tabcolsep}{9pt} 
    \scalebox{0.95}{
    \begin{tabular}{l ccccc ccc} 
    \toprule	 
    \multirow{2}{*}{\textbf{Model}} & \multicolumn{5}{c}{\textbf{NLG Metrics}} & \multicolumn{3}{c}{\textbf{CE Metrics}} \\ 
    \cmidrule(lr){2-6} \cmidrule(lr){7-9}
    & B@1 & B@2 & B@3 & B@4 & R-L & Precision & Recall & F1 \\
    \midrule
    \rowcolor{lightgray}
    \multicolumn{9}{c}{\textbf{\textit{Early Image Captioning Methods}}} \\
    \midrule
    ST \cite{vinyals2015tell}             & 0.299 & 0.184 & 0.121 & 0.084 & 0.263 & 0.249 & 0.203 & 0.204 \\
    Att2In \cite{rennie2017selfcritical}         & 0.325 & 0.203 & 0.136 & 0.096 & 0.276 & 0.322 & 0.239 & 0.249 \\
    AdaAtt \cite{lu2017knowing}          & 0.299 & 0.185 & 0.124 & 0.088 & 0.266 & 0.268 & 0.186 & 0.181 \\
    TopDown \cite{anderson2018bottomup}       & 0.317 & 0.195 & 0.130 & 0.092 & 0.267 & 0.320 & 0.231 & 0.238 \\
    \midrule
    \rowcolor{lightgray}
    \multicolumn{9}{c}{\textbf{\textit{From-Scratch RRG Methods}}} \\
    \midrule
    R2Gen \cite{chen-etal-2020-generating}       & 0.353 & 0.218 & 0.145 & 0.103 & 0.277 & 0.333 & 0.273 & 0.276 \\
    CA \cite{liu2021exploring}           & 0.350 & 0.219 & 0.152 & 0.109 & 0.283 & -     & -     & -     \\
    CMCL \cite{liu-etal-2021-contrastive}       & 0.344 & 0.217 & 0.140 & 0.097 & 0.281 & -     & -     & -     \\
    PPKED \cite{liu2021ppked}            & 0.360 & 0.224 & 0.149 & 0.106 & 0.284 & -     & -     & -     \\
    R2GenCMN \cite{chen-etal-2021-cross-modal}        & 0.353 & 0.218 & 0.148 & 0.106 & 0.278 & 0.334 & 0.275 & 0.278 \\
    R2GenRL \cite{qin-song-2022-reinforced}     & 0.381 & 0.232 & 0.155 & 0.109 & 0.287 & 0.342 & 0.294 & 0.292 \\
    ITA \cite{wang-etal-2022-inclusive}               & 0.395 & 0.253 & 0.170 & 0.121 & 0.284 & -     & -     & -     \\
    WarmStart \cite{nicolson2022warmstart} & 0.392 & 0.245 & 0.169 & 0.124 & 0.285 & 0.359 & 0.412 & 0.384 \\
    KiUT \cite{huang2023kiut} & 0.393 & 0.243 & 0.159 & 0.113 & 0.285 & 0.371 & 0.318 & 0.321 \\
    PromptMRG \cite{jin2024promptmrg} & 0.398 & - & - & 0.112 & 0.258 & 0.501 & 0.509 & 0.476 \\
    RGRG \cite{tanida2023interactive} & 0.373 & 0.249 & \underline{0.175} & 0.126 & 0.264 & 0.461 & 0.475 & 0.447 \\
    \midrule
    \rowcolor{lightgray}
    \multicolumn{9}{c}{\textbf{\textit{Large Language Model-based RRG Methods}}} \\
    \midrule
    XrayGPT \cite{thawkar2023xraygpt} & 0.128 & 0.045 & 0.014 & 0.004 & 0.111 & - & - & - \\
    Med-PaLM \cite{tu2023medpalm} & 0.317 & - & - & 0.115 & 0.275 & - & - & 0.378 \\
    R2GenGPT \cite{wang2023r2gengpt} & 0.396 & - & - & 0.113 & 0.273 & 0.506 & 0.414 & 0.456 \\
    EKAGen \cite{bu2024instance} & 0.419 & \underline{0.258} & 0.170 & 0.119 & 0.287 & 0.517 & 0.483 & 0.499 \\
    CheXAgent \cite{chen2024chexagent} & 0.189 & - & - & 0.040 & 0.208 & 0.506 & 0.306 & 0.381 \\
    MAIRA-1 \cite{hyland2023maira1} & 0.392 & - & - & \underline{0.142} & 0.289 & - & - & 0.553 \\
    R2-LLM \cite{liu2024bootstrapping} & 0.402 & - & - & 0.128 & 0.291 & 0.465 & 0.482 & 0.473 \\
    InVERGe \cite{deria2024inverge} & \textbf{0.425} & 0.240 & 0.132 & 0.100 & 0.309 & - & - & - \\
    LLM-RG4 \cite{wang-etal-2025-llm-rg4} & 0.377 & - & - & 0.144 & \underline{0.318} & \underline{0.583} & \underline{0.593} & \underline{0.588} \\
    \midrule
    \textbf{\textsc{S2D-Align}} (Ours) & \underline{0.422} & \textbf{0.263} & \textbf{0.183} & \textbf{0.149} & \textbf{0.332} & \textbf{0.613} & \textbf{0.606} & \textbf{0.608} \\
    \bottomrule
    \end{tabular}}
    \vspace{-0.7em}
    \caption{Comparison with state-of-the-art methods on the \textsc{MIMIC-CXR} benchmark~\cite{johnson2019mimic} with respect to standard NLG and CE metrics. The best and second best results are highlighted in \textbf{bold} and \underline{underlined}, respectively.}
    \vspace{-2.2em}
    \label{tab:sota_comparison_mimic}
    \end{table*}

\paragraph{Stage 3: Fine-grained Key Phrase Grounding.}
With solid instance-level visual understanding, this final stage aims to explicitly steer the model towards anatomically-grounded alignment.
We first extract a set of clinically-relevant key phrases $K = \{k_1, k_2, \dots, k_m\}$ from the ground-truth report $R_{gt}$ using an entity extraction tool RadGraph \cite{jain2021radgraph}. 
To ensure the grammatical coherence of the extracted entities, we first compose them into a short description if its corresponding relation is positive, then adopt an LLM to refine it into a more natural and clinically-relevant phrase.\footnote{Appendix A shows the used refinement prompt.}
With all training samples annotated, we randomly sample $l$ key phrases from $K$ and feed them to the LLM $\mathcal{G}_{\text{LLM}}$ for anotminal grounding.
Particularly, we use the same medical text encoder as that in last stage to convert the key phrases into the corresponding representation $E_{\text{key}} \in \mathbb{R}^{l \times D_t}$ and use another SMA ($\text{SMA}_p$ with parameters $\Theta_{\text{SMA}_p}$) to map $E_{\text{key}}$ into the same feature space as $V$ and $E_{\text{ref}}$.
Finally, we send the concatenation of $V_{\text{mem}}$, $E_{\text{ref}}$, and $E_{\text{key}}$ to the LLM $\mathcal{G}_{\text{LLM}}$, with the training objective $\mathcal{L}^3_{\text{PAG}}$ formulated by:
\vspace{-0.5em}
\begin{equation} \label{eq:stage_3_loss}
\vspace{-0.5em}
\begin{aligned}
\mathcal{L}^3_{\text{PAG}} =& - \sum_{t=1}^{|R_{gt}|} \log p(w_t | w_{<t}, V_{\text{mem}}, E_{\text{ref}}, E_{\text{key}}; \\
&\Theta_{\text{SMA}_v}, \Theta_{\text{SMA}_t}, \Theta_{\text{SMA}_p})
\end{aligned}
\end{equation}
By conducting this stage, the model is enforced to establish a more precise correspondence between visual regions and their textual descriptions, which is the cornerstone of trustworthy RRG.
In inference, we discard all aforementioned auxiliary signals and perform RRG similar to standard MLLMs, resulting in the generated report $R$, where our experiments below demonstrate such paradigm effectively assists single radiograph-to-report generation without annotational guidance.

\paragraph{Training Protocol of PAG}
Our training protocol strictly follows the three-stage curriculum in a sequential manner.
Specifically, we first train the model exclusively on the Stage 1 objective until the validation loss converges.
The resulting model weights then serve as the initialization for Stage 2, where we continue the training process by introducing the additional reference report and its corresponding SMA module.
Similar process is repeated for Stage 3 with an integration of key phrases.
By performing this, the sequential paradigm ensures the model to first establish a coarse visual-textual foundation before being tasked with learning finer-grained contextual and anatomical relationships.

\begin{table}[t]
\centering
\caption{Comparison with state-of-the-art methods on \textsc{IU X-Ray} \cite{iu-xray} w.r.t. NLG metrics.}
\vspace{-0.5em}
\label{tab:sota_comparison_iuxray}
\footnotesize
\setlength{\tabcolsep}{5pt}
\scalebox{0.95}{\begin{tabular}{l ccc} 
\toprule	 
Methods & B@1 & B@4 & R-L \\
\midrule
\rowcolor{lightgray}
\multicolumn{4}{c}{\textbf{\textit{Early Image Captioning Methods}}} \\
\midrule
ST \cite{vinyals2015tell}             & 0.216 & 0.066 & 0.306 \\
Att2In \cite{rennie2017selfcritical}         & 0.224 & 0.068 & 0.308 \\
ADAATT \cite{lu2017knowing}          & 0.220 & 0.068 & 0.308 \\
CoATT \cite{jing2018automatic}             & 0.455 & 0.154 & 0.369 \\
HRGR \cite{nips-2018-hrgr-agent}              & 0.438 & 0.151 & 0.322 \\
CMAS-RL \cite{jing-etal-2019-show}        & 0.464 & 0.154 & 0.362 \\
\midrule
\rowcolor{lightgray}
\multicolumn{4}{c}{\textbf{\textit{From-Scratch RRG Methods}}} \\
\midrule
R2Gen \cite{chen-etal-2020-generating}       & 0.470 & 0.165 & 0.371 \\
CA \cite{liu2021exploring}           & 0.492 & 0.169 & 0.381 \\
CMCL \cite{liu-etal-2021-contrastive}       & 0.473 & 0.162 & 0.378 \\
PPKED \cite{liu2021ppked}            & 0.483 & 0.168 & 0.376 \\
R2GenCMN \cite{chen-etal-2021-cross-modal}        & 0.475 & 0.170 & 0.375 \\
R2GenRL \cite{qin-song-2022-reinforced}     & 0.494 & 0.181 & 0.384 \\
\midrule
\rowcolor{lightgray}
\multicolumn{4}{c}{\textbf{\textit{Large Language Model-based RRG Methods}}} \\
\midrule
XrayGPT (7B) \cite{thawkar2023xraygpt} & 0.177 & 0.007 & 0.203 \\
R2-LLM (14.2B)$^{\dagger}$ \cite{liu2024bootstrapping} & \underline{0.499} & \underline{0.184} & \underline{0.390} \\
\midrule
\textbf{Ours (7B)} & \textbf{0.512} & \textbf{0.195} & \textbf{0.407} \\
\bottomrule
\end{tabular}}
\vspace{-2.2em}
\end{table}

\vspace{-0.8em}
\section{Experimental Setup}
\vspace{-0.8em}
\subsection{Datasets}
\vspace{-0.5em}

We conduct our primary experiments on two benchmark datasets, i.e., \textsc{IU X-Ray} and \textsc{MIMIC-CXR}.
\textsc{IU X-Ray} \cite{iu-xray} is collected by the Indiana University, where it serves one of the most widely adopted benchmarks for RRG, containing $7,470$ chest X-ray images and $3,955$ corresponding radiology reports.
\textsc{MIMIC-CXR} \cite{johnson2019mimic} is the largest publicly available dataset of chest X-ray radiographs and their corresponding free-text radiology reports, collected from the Beth Israel Deaconess Medical Center (BIDMC), where it contains $377,110$ image-report pairs from $65,379$ patients.
Following the pre-processing pipeline of conventional studies \cite{chen-etal-2020-generating,chen-etal-2021-cross-modal,liu2024bootstrapping, wang-etal-2025-llm-rg4}, we extract the ``Findings'' section from each report for our analysis.
We adhere to the official data split, consisting of $270,790$ training, $2,130$ validation, and $3,858$ testing samples.
For data samples on \textsc{MIMIC-CXR}, we exclude samples lacking RadGraph annotations \cite{jain2021radgraph} to ensure that all key phrases are properly annotated.

\begin{figure*}[t!]
\centering
\includegraphics[width=0.9\linewidth]{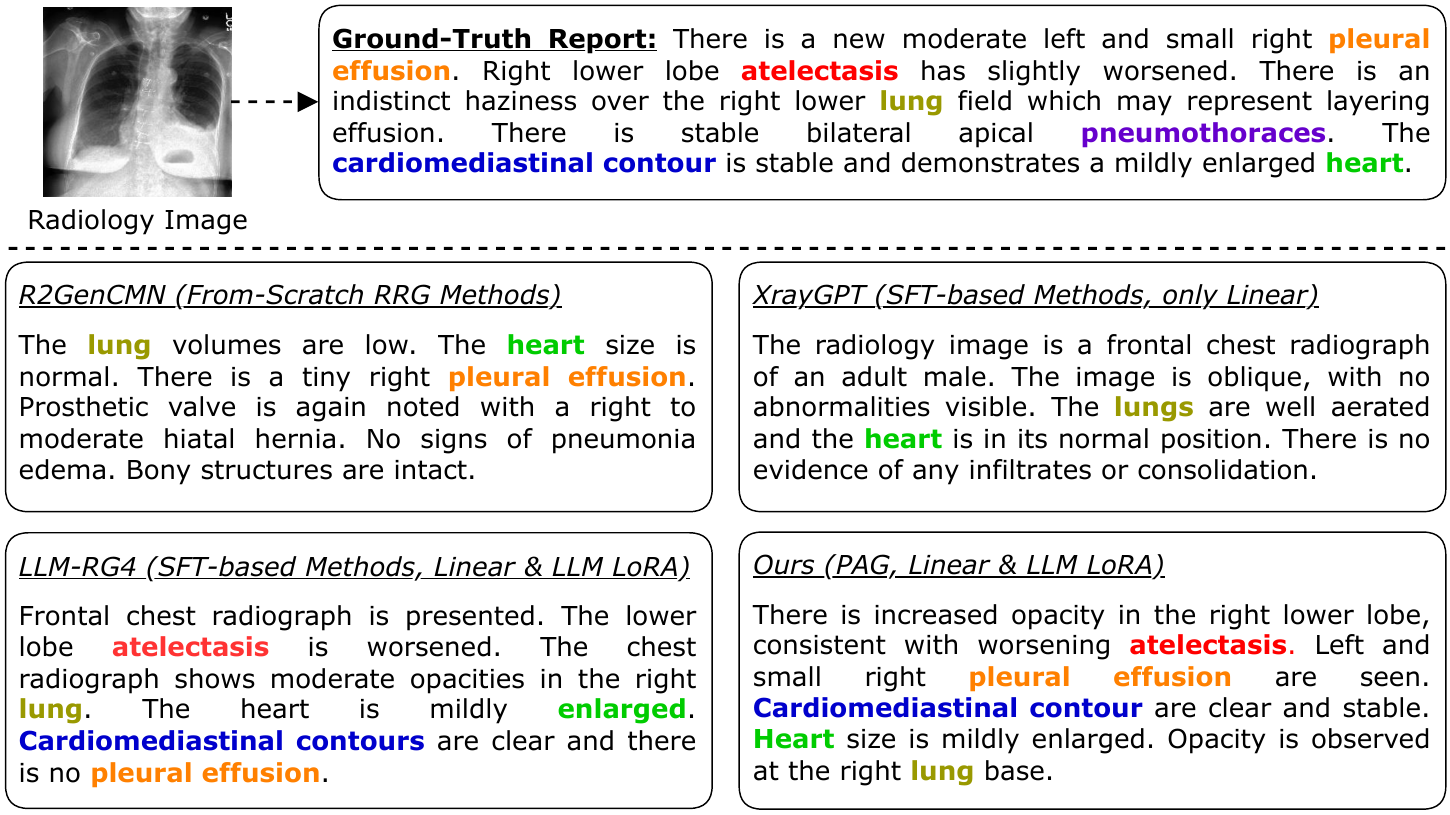}
\vspace{-0.8em}
\caption{\textbf{A case study selected from \textsc{MIMIC-CXR},} with medical concepts shared by the ground-truth and generated outputs highlighted in the same color. 
The categories and optimized parameters for the LLM-based methods are detailed in parentheses.
}
\vspace{-1.8em}
\label{fig:case_study}
\end{figure*}

\vspace{-0.5em}
\subsection{Evaluation Metrics}
\vspace{-0.3em}
Following standard practice~\cite{nips-2018-hrgr-agent,chen2020r2gen,chen2021r2cmn,qin2022reinforced}, we first evaluate the generated reports using established metrics for Natural Language Generation (NLG) and Clinical Efficacy (CE).
Specifically, we employ BLEU-$n$~\cite{papineni2002bleu} (B@$n$, $n \in \{1, 4\}$) and ROUGE-L~\cite{lin2004rouge} (R-L) for NLG assessment, and report precision, recall, and F1 scores for CE evaluation.
Additionally, we further conduct a human evaluation to assess report quality from a clinical perspective\footnote{Appendix B reports the human evaluation results.}.
%

\vspace{-0.2em}
\subsection{Implementation Details}
\vspace{-0.2em}
For our implementation of the visual encoder $\mathcal{E}_v$, we leverage the pre-trained Rad-DINO \cite{perez2024raddino}, and keep its parameters frozen throughout the fine-tuning process, in order to preserve its domain-specific visual representations learned from large-scale medical data.
For the LLM decoder, we initialize the first stage of PAG with \texttt{Vicuna-7B-v1.5}, whose parameters are fixed during the first PAG stage, and fine-tuned via Low-Rank Adaptation (LoRA) \cite{hu2021lowrank}.
Herein, we apply LoRA to all linear layers of the transformer blocks with a rank of $16$, a scaling factor of $16$, and a dropout rate of $0.1$, where the original LLM parameters are kept frozen, with only the LoRA adapter weights being updated during training.
For the text encoder $\mathcal{E}_{\text{text}}$, we use \texttt{BiomedVLP-CXR-BERT-specialized} to encode the reference report and key phrases.
For the SMA in different stages of PAG, it consists of a 8-head cross-attention layer followed by a 3-layer MLP and Layer Normalization (LN).
The hyperparameters for \textsc{S2D-Align} were tuned on the validation set, where we performed a grid search over key parameters to identify the optimal configuration\footnote{Appendix C details our hyper-parameter settings.}.

\vspace{-0.2em}
\section{Results and Analysis}

\paragraph{Comparison with State-of-the-Art Methods.}
As presented in the quantitative comparison of Table \ref{tab:sota_comparison_mimic} and \ref{tab:sota_comparison_iuxray}, we conduct a comprehensive comparison of our proposed \textsc{S2D-Align} with a wide range of state-of-the-art methods on both the \textsc{IU X-Ray} \cite{iu-xray} and \textsc{MIMIC-CXR} \cite{johnson2019mimic}.
Our proposed \textsc{S2D-Align} establishes a new state-of-the-art on both benchmarks, achieving a F1 score of $0.608$ on the CE metric.
This superiority stems directly from our PAG strategy to progressively guide the model towards fine-grained radiograph-report alignment, which shows significantly lower performance on factual correctness (i.e., CE and RadGraph-based metrics) and reveals their failure to precisely map visual findings to precise textual descriptions.
In contrast, the leading performance demonstrated by \textsc{S2D-Align} provide direct evidence that it establishes \textit{anatomically-grounded alignment} crucial for clinical reliability.
Furthermore, our end-to-end paradigm, unified by the SMA design, overcomes the limitations of ICL (e.g., Yan et al. \cite{yan2023style}) by avoiding reliance on potentially noisy intermediate textual representations, meanwhile enabling the models to discover more potential complementary information through feature sharing across modalities.
These results validate that \textsc{S2D-Align} is a more robust and factually reliable direction.


\paragraph{Case Study.}
We present a qualitative case study to intuitively compare the capabilities of representative RRG methods in Figure~\ref{fig:case_study}, with models from three categories, i.e., a from-scratch RRG method (R2GenCMN), an SFT-based MLLM with only the projector trained (XrayGPT), and an SFT-based MLLM that updates parameters of the LLM (LLM-RG4).
As presented, R2GenCMN fails to capture key abnormalities like ``atelectasis'' and hallucinates non-existent findings such as a ``prosthetic valve'', showcasing the limitations of training without leveraging pre-trained knowledge.
XrayGPT produces a coarse-grained report with facutally incorrect diagnosis, i.e., ``no abnormalities visible'', and misses most pathological findings, since the simple MLP project fails to model the complex visual-textual mapping.
While updating the LLM parameters allows LLM-RG4 to correctly identify more findings like ``atelectasis'', the fundamental issue of standard SFT leads to its coarse-grained alignment, which explicitly denies the presence of ``pleural effusion'' that is clearly visible in the image.
All aforementioned issues are alleviated by \textsc{S2D-Align} with comprehensive findings, meanwhile maintaining high descriptive quality.
This vividly demonstrates that simply fine-tuning the LLM is insufficient to obtain an anatomically-grounded alignment, where PAG is crucial to empower the model to move beyond coarse pattern matching and perform factually-correct clinical reasoning.

\vspace{-0.8em}
\section{Ablation Studies}
\vspace{-0.5em}

\begin{table}[t!]
    \centering
\caption{Evaluation scores for the ablation studies of PAG on \textsc{MIMIC-CXR}, where ``Si'' denotes Stage \textit{i}, ``$\rightarrow$'' indicates progressive training, and ``+'' indicates joint training.}
\vspace{-0.5em}
\label{tab:pag_ablation}
\setlength{\tabcolsep}{4.5pt}
\resizebox{\columnwidth}{!}{%
\begin{tabular}{l|cccc}
\toprule
\textbf{Method} & \textbf{B@1} & \textbf{B@4} & \textbf{R-L} & \textbf{F1} \\
\midrule
\rowcolor{lightgray}
\multicolumn{5}{l}{\textit{Evaluating the progressive nature of PAG}} \\
\midrule
Single-Stage (S1 only) & 0.369 & 0.092 & 0.271 & 0.449 \\
Joint Training (S1+S2+S3) & 0.371 & 0.102 & 0.295 & 0.470 \\
Reversed Order (S1$\rightarrow$S3$\rightarrow$S2) & 0.355 & 0.077 & 0.301 & 0.527 \\
\midrule
\rowcolor{lightgray}
\multicolumn{5}{l}{\textit{Evaluating the contribution of each PAG stage}} \\
\midrule
w/o Fine-grained Grounding (S1$\rightarrow$S2) & 0.386 & 0.112 & 0.303 & 0.513 \\
w/o Contextual Enhancement (S1$\rightarrow$S3) & \underline{0.415} & \underline{0.145} & \underline{0.318} & \underline{0.565} \\
\midrule
\textbf{S2D-Align (Full, S1$\rightarrow$S2$\rightarrow$S3)} & \textbf{0.422} & \textbf{0.149} & \textbf{0.332} & \textbf{0.608} \\
\bottomrule
\end{tabular}%
}
\vspace{-1.5em}
\end{table}

\subsection{Effect of PAG}
To validate the design of PAG, we conduct extensive ablation studies by exploring different training paradigms, with results detailed in Table~\ref{tab:pag_ablation}.
We consider three variants of training paradigms, i.e., training with only the coarse-grained stage (``Single-Stage''), mixing all data for ``Joint Training'', and training in a ``Reversed Order'' (S1$\rightarrow$S3$\rightarrow$S2), along with two baselines ablating the second and third training stages of PAG.
It is observed that ``Single-Stage'' and ``Joint Training'' result in substantially lower performance, particularly on the F1 score that reflects the factual correctness, indicating that standard SFT (radiograph-report alignment) is insufficient while mixed training struggles to guide the model to learn a coarse-to-fine alignment process.
Similar results are seen in ``S1 $\rightarrow$ S2'' where fine-grained grounding is removed, showing that injecting key phrases as an auxiliary signal is crucial for the shallow-to-deep learning.
Interestingly, by comparing ``S1 $\rightarrow$ S3'' and ``S1 $\rightarrow$ S3 $\rightarrow$ S2'', we observe that fine-tuning in a reversed order might harm the performance of directly aligning with key phrases, which underscores a potential impact of PAG similar to that of curriculum learning, with the best results demonstrated by our full model.
This suggests that a carefully structured curriculum for domain-specific MLLM, which first establishes a foundational understanding, and then refines it with increasingly granular supervision, is a more effective than simply exposing it to massive data.

\begin{table}[t!]
\centering
\caption{Evaluation scores for the ablation studies of SMA on \textsc{MIMIC-CXR}, where ``MLP'' and ``MLP + Q-Former'' denote the linear projection and its combination with Q-Former \cite{li2023blip2}.
``SMA (MLP + MSA)'' and ``SMA (w/o Shared Memory)'' indicate the architecture without the memory bank and the sharing mechanism, respectively.}
\vspace{-0.5em}
\label{tab:sma_ablation}
\small
\setlength{\tabcolsep}{5pt} 
\begin{tabular}{l|cccc}
\toprule
\textbf{Connector Module} & B@1 & B@4 & R-L & F1 \\
\midrule
MLP & 0.387 & 0.104 & 0.283 & 0.473 \\
MLP + Q-Former & 0.368 & 0.098 & 0.264 & 0.394 \\
SMA (MLP + MSA) & \underline{0.407} & 0.119 & 0.311 & 0.523 \\
SMA (w/o Shared Memory) & 0.401 & \underline{0.136} & \underline{0.320} & \underline{0.559} \\
\midrule
\textbf{SMA (Ours)} & \textbf{0.422} & \textbf{0.149} & \textbf{0.332} & \textbf{0.608} \\
\bottomrule
\end{tabular}
\vspace{-2em}
\end{table}

\vspace{-0.6em}
\subsection{Effect of SMA}
\vspace{-0.2em}
To investigate the effect of SMA, we replace it with other variants of connector modules, and ablate the SMA design, with results summarized in Table~\ref{tab:sma_ablation}.
Both the MLP and the MLP with a Q-Former yield drastically lower scores across all metrics, particularly on F1 score, which indicate that conventional connectors struggle to model complex and fine-grained visual-textual mappings for the purpose of RRG.
Notably in this comparison, the integration of Q-Former causes further performance degradation, since Q-Former might lead to possible information loss due to feature compression, where this conclusion is consistent with some up-to-date MLLM studies \cite{lin-etal-2023-sphinx}.
While the basic SMA architecture improves the performance owing to increased model parameters (MLP + MSA) or the memory mechanism (SMA w/o Shared Memory), the lack of feature sharing still prevents the model from achieving further anatomically-grounded alignment, where the best results are achieved by our full SMA design, confirming the effectiveness of the feature sharing across different stages of PAG.

\vspace{-0.6em}
\section{Conclusion}
\vspace{-0.2em}
In this paper, we addressed the critical challenge of factual correctness in RRG, which is undermined by the coarse-grained alignment in standard SFT-based fine-tuning methods.
To this end, we introduced \textsc{S2D-Align}, a novel paradigm centered on our PAG strategy, which employs a shallow-to-deep curriculum to explicitly establish anatomically-grounded alignment.
This multi-stage process is effectively unified by our lightweight SMA, designed to integrate multi-granularity guidance and overcome the limitations of simple projection layers.
Our comprehensive experiments demonstrate that \textsc{S2D-Align} sets a new state-of-the-art on \textsc{IU X-Ray} and \textsc{MIMIC-CXR}, significantly enhancing the factual reliability and clinical utility of generated reports.
Ultimately, this work validates that pursuing anatomically-grounded learning is a pivotal direction for building more trustworthy generative models, where we wish it can serve as a reference work for follow-up studies.

\bibliography{aaai2026}

@String{Computer = "{IEEE} Computer" }

@String(CVPR= {IEEE Conf. Comput. Vis. Pattern Recog.})

@String(ICLR = {Int. Conf. Learn. Represent.})

@String(AAAI = {AAAI})

@String(CVPR  = {CVPR})

@String(ICLR  = {ICLR})

@article{jain2021radgraph,
  title={RadGraph: Extracting Clinical Entities and Relations from Radiology Reports},
  author={Jain, Saahil and Agrawal, Ashwin and Saporta, Adriel and Truong, Steven QH and Nguyen Duong, Du and Bui, Tan and Chambon, Pierre and Zhang, Yuhao and Lungren, Matthew P and Ng, Andrew Y and others},
  journal={arXiv preprint arXiv:2106.14463},
  year={2021}
}

@article{touvron2023llama,
  title={Llama: Open and Efficient Foundation Language Models},
  author={Touvron, Hugo and Lavril, Thibaut and Izacard, Gautier and Martinet, Xavier and Lachaux, Marie-Anne and Lacroix, Timoth{\'e}e and Rozi{\`e}re, Baptiste and Goyal, Naman and Hambro, Eric and Azhar, Faisal and others},
  journal={arXiv preprint arXiv:2302.13971},
  year={2023}
}

@inproceedings{jin2024promptmrg,
  title={{PromptMRG: Diagnosis-driven Prompts for Medical Report Generation}},
  author={Jin, Haibo and Che, Haoxuan and Lin, Yi and Chen, Hao},
  booktitle={Proceedings of the AAAI Conference on Artificial Intelligence},
  volume={38},
  number={3},
  pages={2607--2615},
  year={2024}
}

@article{hyland2023maira1,
  author       = {Stephanie L. Hyland and
                  Shruthi Bannur and
                  Kenza Bouzid and
                  Daniel C. Castro and
                  Mercy Ranjit and
                  Anton Schwaighofer and
                  Fernando P{\'{e}}rez{-}Garc{\'{\i}}a and
                  Valentina Salvatelli and
                  Shaury Srivastav and
                  Anja Thieme and
                  Noel Codella and
                  Matthew P. Lungren and
                  Maria Teodora Wetscherek and
                  Ozan Oktay and
                  Javier Alvarez{-}Valle},
  title        = {{MAIRA-1:} {A} Specialised Large Multimodal Model for Radiology Report
                  Generation},
  journal      = {CoRR},
  volume       = {abs/2311.13668},
  year         = {2023},
}

@article{johnson2019mimic,
  title={MIMIC-CXR, A De-identified Publicly Available Database of Chest Radiographs with free-text reports},
  author={Johnson, Alistair EW and Pollard, Tom J and Berkowitz, Seth J and Greenbaum, Nathaniel R and Lungren, Matthew P and Deng, Chih-ying and Mark, Roger G and Horng, Steven},
  journal={Scientific data},
  volume={6},
  number={1},
  pages={317},
  year={2019},
}

@inproceedings{papineni2002bleu,
  title={Bleu: A Method for Automatic Evaluation of Machine Translation},
  author={Papineni, Kishore and Roukos, Salim and Ward, Todd and Zhu, Wei-Jing},
  booktitle={Proceedings of the 40th annual meeting of the Association for Computational Linguistics},
  pages={311--318},
  year={2002}
}

@inproceedings{lin2004rouge,
  title={Rouge: A Package for Automatic Evaluation of Summaries},
  author={Lin, Chin-Yew},
  booktitle={Text summarization branches out},
  pages={74--81},
  year={2004}
}

@article{perez2024raddino,
  title={Rad-dino: Exploring Scalable Medical Image Encoders Beyond Text Supervision},
  author={P{\'e}rez-Garc{\'\i}a, Fernando and Sharma, Harshita and Bond-Taylor, Sam and Bouzid, Kenza and Salvatelli, Valentina and Ilse, Maximilian and Bannur, Shruthi and Castro, Daniel C and Schwaighofer, Anton and Lungren, Matthew P and others},
  journal={arXiv preprint arXiv:2401.10815},
  year={2024}
}

@inproceedings{liu2021ppked,
  title={Exploring and Distilling Posterior and Prior Knowledge for Radiology Report Generation},
  author={Liu, Fenglin and Wu, Xian and Ge, Shen and Fan, Wei and Zou, Yuexian},
  booktitle={Proceedings of the IEEE/CVF conference on computer vision and pattern recognition},
  pages={13753--13762},
  year={2021}
}

@inproceedings{yan2023style,
    title = "Style-Aware Radiology Report Generation with {R}ad{G}raph and Few-Shot Prompting",
    author = "Yan, Benjamin  and
      Liu, Ruochen  and
      Kuo, David  and
      Adithan, Subathra  and
      Reis, Eduardo  and
      Kwak, Stephen  and
      Venugopal, Vasantha  and
      O{'}Connell, Chloe  and
      Saenz, Agustina  and
      Rajpurkar, Pranav  and
      Moor, Michael",
    booktitle = "Findings of the Association for Computational Linguistics: EMNLP 2023",
    year = "2023",
}

@inproceedings{bu2024instance,
  title={Instance-level Expert Knowledge and Aggregate Discriminative Attention for Radiology Report Generation},
  author={Bu, Shenshen and Li, Taiji and Yang, Yuedong and Dai, Zhiming},
  booktitle={Proceedings of the IEEE/CVF Conference on Computer Vision and Pattern Recognition},
  pages={14194--14204},
  year={2024}
}

@inproceedings{chen2020r2gen,
  author       = {Zhihong Chen and
                  Yan Song and
                  Tsung{-}Hui Chang and
                  Xiang Wan},
  title        = {Generating Radiology Reports via Memory-driven Transformer},
  booktitle    = {EMNLP},
  pages        = {1439--1449},
  year         = {2020},
}

@inproceedings{chen2021r2cmn,
  author       = {Zhihong Chen and
                  Yaling Shen and
                  Yan Song and
                  Xiang Wan},
  title        = {{Cross-modal Memory Networks for Radiology Report Generation}},
  booktitle    = {ACL},
  pages        = {5904--5914},
  year         = {2021},
  }

@inproceedings{qin2022reinforced,
  author       = {Han Qin and
                  Yan Song},
  title        = {{Reinforced Cross-modal Alignment for Radiology Report Generation}},
  booktitle    = {ACL},
  pages        = {448--458},
  year         = {2022},
}

@article{tu2023medpalm,
  author       = {Tao Tu and
                  Shekoofeh Azizi and
                  Danny Driess and
                  Mike Schaekermann and
                  Mohamed Amin and
                  Pi{-}Chuan Chang and
                  Andrew Carroll and
                  Chuck Lau and
                  Ryutaro Tanno and
                  Ira Ktena and
                  Basil Mustafa and
                  Aakanksha Chowdhery and
                  Yun Liu and
                  Simon Kornblith and
                  David J. Fleet and
                  Philip Andrew Mansfield and
                  Sushant Prakash and
                  Renee Wong and
                  Sunny Virmani and
                  Christopher Semturs and
                  S. Sara Mahdavi and
                  Bradley Green and
                  Ewa Dominowska and
                  Blaise Ag{\"{u}}era y Arcas and
                  Joelle K. Barral and
                  Dale R. Webster and
                  Gregory S. Corrado and
                  Yossi Matias and
                  Karan Singhal and
                  Pete Florence and
                  Alan Karthikesalingam and
                  Vivek Natarajan},
  title        = {{Towards Generalist Biomedical {AI}}},
  journal      = {CoRR},
  volume       = {abs/2307.14334},
  year         = {2023},
  pages = {},
}

@inproceedings{jing2018automatic,
    title = "{On the Automatic Generation of Medical Imaging Reports}",
    author = "Jing, Baoyu  and
      Xie, Pengtao  and
      Xing, Eric",
    booktitle = "ACL 2018",
    month = jul,
    year = "2018",
    address = "Melbourne, Australia",
    pages = "2577--2586",
}

@inproceedings{tanida2023interactive,
    title={{Interactive and Explainable Region-guided Radiology Report Generation}},
    author={Tanida, Tim and Müller, Philip and Kaissis, Georgios and Rueckert, Daniel},
    booktitle={CVPR},
    year={2023}
}

@inproceedings{nips-2018-hrgr-agent,
  author       = {Yuan Li and
                  Xiaodan Liang and
                  Zhiting Hu and
                  Eric P. Xing},
  title        = {{Hybrid Retrieval-Generation Reinforced Agent for Medical Image Report
                  Generation}},
  booktitle    = {NeurIPS 2018},
  pages        = {1537--1547},
  year         = {2018},
  bibsource    = {dblp computer science bibliography, https://dblp.org}
}

@inproceedings{liu2024bootstrapping,
  author       = {Chang Liu and
                  Yuanhe Tian and
                  Weidong Chen and
                  Yan Song and
                  Yongdong Zhang},
  title        = {{Bootstrapping Large Language Models for Radiology Report Generation}},
  booktitle    = {AAAI},
  pages        = {18635--18643},
  year         = {2024},
}

@inproceedings{jing-etal-2019-show,
    title = {{Show, Describe and Conclude: On Exploiting the Structure Information of Chest {X}-ray Reports}},
    author = {Baoyu Jing and Zeya Wang and Eric Xing},
    booktitle = "ACL 2019",
    month = jul,
    year = "2019",
    address = "Florence, Italy",
    pages = "6570--6580",
}

@inproceedings{jing-etal-2018-automatic,
    title = "{O}n the {A}utomatic {G}eneration of {M}edical {I}maging {R}eports",
    author = {Baoyu Jing and Pengtao Xie and Eric Xing},
    booktitle = "Proceedings of the 56th Annual Meeting of the Association for Computational Linguistics (Volume 1: Long Papers)",
    month = jul,
    year = "2018",
    address = "Melbourne, Australia",
    pages = "2577--2586",
}

@misc{thawkar2023xraygpt,
      title={{X}ray{GPT}: {C}hest {R}adiographs {S}ummarization using {M}edical {V}ision-{L}anguage {M}odels}, 
      author={Omkar Thawkar and Abdelrahman Shaker and Sahal Shaji Mullappilly and Hisham Cholakkal and Rao Muhammad Anwer and Salman Khan and Jorma Laaksonen and Fahad Shahbaz Khan},
      year={2023},
      eprint={2306.07971},
      archivePrefix={arXiv},
      primaryClass={cs.CV}
}

@inproceedings{chen-etal-2020-generating,
    title = "Generating Radiology Reports via Memory-driven Transformer",
    author = "Chen, Zhihong  and
      Song, Yan  and
      Chang, Tsung-Hui  and
      Wan, Xiang",
    booktitle = "Proceedings of the 2020 Conference on Empirical Methods in Natural Language Processing (EMNLP)",
    month = nov,
    year = "2020",
    address = "Online",
    pages = "1439--1449",
}

@inproceedings{chen-etal-2021-cross-modal,
    title = "Cross-modal Memory Networks for Radiology Report Generation",
    author = "Chen, Zhihong  and
      Shen, Yaling  and
      Song, Yan  and
      Wan, Xiang",
    booktitle = "Proceedings of the 59th Annual Meeting of the Association for Computational Linguistics and the 11th International Joint Conference on Natural Language Processing (Volume 1: Long Papers)",
    month = aug,
    year = "2021",
    address = "Online",
    pages = "5904--5914",
}

@inproceedings{qin-song-2022-reinforced,
    title = "Reinforced Cross-modal Alignment for Radiology Report Generation",
    author = "Qin, Han  and
      Song, Yan",
    booktitle = "Findings of the Association for Computational Linguistics: ACL 2022",
    month = may,
    year = "2022",
    address = "Dublin, Ireland",
    pages = "448--458",
}

@misc{hu2021lowrank,
  author = {Hu, Edward J. and Shen, Yelong and Wallis, Phillip and Allen-Zhu, Zeyuan and Li, Yuanzhi and Wang, Shean and Wang, Lu and Chen, Weizhu},
  description = {[2106.09685] LoRA: Low-Rank Adaptation of Large Language Models},
  note = {cite arxiv:2106.09685Comment: Draft V2 includes better baselines, experiments on GLUE, and more on  adapter latency},
  title = {LoRA: Low-Rank Adaptation of Large Language Models},
  year = 2021
}

@misc{li2023blip2,
      title={BLIP-2: Bootstrapping Language-image Pre-training with Frozen Image Encoders and Large Language Models}, 
      author={Junnan Li and Dongxu Li and Silvio Savarese and Steven Hoi},
      year={2023},
      eprint={2301.12597},
      archivePrefix={arXiv},
      primaryClass={cs.CV}
}

@misc{zhu2023minigpt4,
      title={MiniGPT-4: Enhancing Vision-language Understanding with Advanced Large Language Models}, 
      author={Deyao Zhu and Jun Chen and Xiaoqian Shen and Xiang Li and Mohamed Elhoseiny},
      year={2023},
      eprint={2304.10592},
      archivePrefix={arXiv},
      primaryClass={cs.CV}
}

@misc{vinyals2015tell,
      title={Show and Tell: A Neural Image Caption Generator}, 
      author={Oriol Vinyals and Alexander Toshev and Samy Bengio and Dumitru Erhan},
      year={2015},
      eprint={1411.4555},
      archivePrefix={arXiv},
      primaryClass={cs.CV}
}

@misc{rennie2017selfcritical,
      title={Self-critical Sequence Training for Image Captioning}, 
      author={Steven J. Rennie and Etienne Marcheret and Youssef Mroueh and Jarret Ross and Vaibhava Goel},
      year={2017},
      eprint={1612.00563},
      archivePrefix={arXiv},
      primaryClass={cs.LG}
}

@misc{lu2017knowing,
      title={Knowing When to Look: Adaptive Attention via A Visual Sentinel for Image Captioning}, 
      author={Jiasen Lu and Caiming Xiong and Devi Parikh and Richard Socher},
      year={2017},
      eprint={1612.01887},
      archivePrefix={arXiv},
      primaryClass={cs.CV}
}

@inproceedings{liu-etal-2021-contrastive,
    title = "Contrastive Attention for Automatic Chest {X}-ray Report Generation",
    author = "Liu, Fenglin  and
      Yin, Changchang  and
      Wu, Xian  and
      Ge, Shen  and
      Zhang, Ping  and
      Sun, Xu",
    booktitle = "Findings of the Association for Computational Linguistics: ACL-IJCNLP 2021",
    month = aug,
    year = "2021",
    address = "Online",
    pages = "269--280",
}

@misc{liu2021exploring,
      title={Exploring and Distilling Posterior and Prior Knowledge for Radiology Report Generation}, 
      author={Fenglin Liu and Xian Wu and Shen Ge and Wei Fan and Yuexian Zou},
      year={2021},
      eprint={2106.06963},
      archivePrefix={arXiv},
      primaryClass={cs.CV}
}

@inproceedings{anderson2018bottomup,
  author = {Anderson, Peter and He, Xiaodong and Buehler, Chris and Teney, Damien and Johnson, Mark and Gould, Stephen and Zhang, Lei},
  booktitle = {CVPR},
  pages = {6077-6086},
  title = {Bottom-up and Top-down Attention for Image Captioning and Visual Question Answering.},
  year = 2018
}

@inproceedings{huang2023kiut,
  author       = {Zhongzhen Huang and
                  Xiaofan Zhang and
                  Shaoting Zhang},
  title        = {{KiUT: Knowledge-injected U-Transformer for Radiology Report Generation}},
  booktitle    = {CVPR},
  pages        = {19809--19818},
  year         = {2023},
}

@InProceedings{wang-etal-2022-inclusive,
author="Wang, Lin and Ning, Munan and Lu, Donghuan and Wei, Dong and Zheng, Yefeng and Chen, Jie",
title="An Inclusive Task-aware Framework for Radiology Report Generation",
booktitle="MICCAI",
year="2022",
}

@misc{nicolson2022warmstart,
      title={Improving Chest X-Ray Report Generation by Leveraging Warm-Starting}, 
      author={Aaron Nicolson and Jason Dowling and Bevan Koopman},
      year={2022},
      eprint={2201.09405},
      archivePrefix={arXiv},
      primaryClass={cs.CV}
}

@article{wang2023r2gengpt,
  author       = {Zhanyu Wang and
                  Lingqiao Liu and
                  Lei Wang and
                  Luping Zhou},
  title        = {{R2GenGPT: Radiology Report Generation with Frozen LLMs}},
  journal      = {arXiv},
  year         = {2023},
}

@article{chen2024chexagent,
  author       = {Zhihong Chen and
                  Maya Varma and
                  Jean{-}Benoit Delbrouck and
                  Magdalini Paschali and
                  Louis Blankemeier and
                  Dave Van Veen and
                  Jeya Maria Jose Valanarasu and
                  Alaa Youssef and
                  Joseph Paul Cohen and
                  Eduardo Pontes Reis and
                  Emily Bao Tsai and
                  Andrew Johnston and
                  Cameron Olsen and
                  Tanishq Mathew Abraham and
                  Sergios Gatidis and
                  Akshay S. Chaudhari and
                  Curtis P. Langlotz},
  title        = {{CheXagent: Towards a Foundation Model for Chest X-Ray Interpretation}},
  journal      = {CoRR},
  year         = {2024},
}

@inproceedings{deria2024inverge,
  author       = {Ankan Deria and
                  Komal Kumar and
                  Snehashis Chakraborty and
                  Dwarikanath Mahapatra and
                  Sudipta Roy},
  title        = {{InVERGe: Intelligent Visual Encoder for Bridging Modalities in Report Generation}},
  booktitle    = {CVPR},
  pages        = {2028--2038},
  year         = {2024},
}

@inproceedings{wang-etal-2025-llm-rg4,
  author       = {Zhuhao Wang and
                  Yihua Sun and
                  Zihan Li and
                  Xuan Yang and
                  Fang Chen and
                  Hongen Liao},
  title        = {{LLM-RG4: Flexible and Factual Radiology Report Generation Across Diverse Input Contexts}},
  booktitle    = {AAAI},
  pages        = {8250--8258},
  year         = {2025},
}

@article{iu-xray,
  author       = {Dina Demner{-}Fushman and
                  Marc D. Kohli and
                  Marc B. Rosenman and
                  Sonya E. Shooshan and
                  Laritza Rodriguez and
                  Sameer K. Antani and
                  George R. Thoma and
                  Clement J. McDonald},
  title        = {{Preparing A Collection of Radiology Examinations for Distribution and Retrieval}},
  journal      = {J. Am. Medical Informatics Assoc.},
  volume       = {23},
  number       = {2},
  pages        = {304--310},
  year         = {2016},
}

@article{lin-etal-2023-sphinx,
  author       = {Ziyi Lin and
                  Chris Liu and
                  Renrui Zhang and
                  Peng Gao and
                  Longtian Qiu and
                  Han Xiao and
                  Han Qiu and
                  Chen Lin and
                  Wenqi Shao and
                  Keqin Chen and
                  Jiaming Han and
                  Siyuan Huang and
                  Yichi Zhang and
                  Xuming He and
                  Hongsheng Li and
                  Yu Qiao},
  title        = {{{SPHINX:} The Joint Mixing of Weights, Tasks, and Visual Embeddings
                  for Multi-modal Large Language Models}},
  journal      = {CoRR},
  year         = {2023},
}

@inproceedings{liu-etal-2023-llava,
  author       = {Haotian Liu and
                  Chunyuan Li and
                  Qingyang Wu and
                  Yong Jae Lee},
  title        = {{Visual Instruction Tuning}},
  booktitle    = {NeurIPS},
  year         = {2023},
}

@inproceedings{dosovitskiy-etal-2021-vit,
  author       = {Alexey Dosovitskiy and
                  Lucas Beyer and
                  Alexander Kolesnikov and
                  Dirk Weissenborn and
                  Xiaohua Zhai and
                  Thomas Unterthiner and
                  Mostafa Dehghani and
                  Matthias Minderer and
                  Georg Heigold and
                  Sylvain Gelly and
                  Jakob Uszkoreit and
                  Neil Houlsby},
  title        = {{An Image is Worth 16x16 Words: Transformers for Image Recognition
                  at Scale}},
  booktitle    = {ICLR},
  year         = {2021},
}

@inproceedings{Bengio2009CurriculumL,
author = {Bengio, Yoshua and Louradour, J\'{e}r\^{o}me and Collobert, Ronan and Weston, Jason},
title = {{Curriculum Learning}},
year = {2009},
isbn = {9781605585161},
address = {New York, NY, USA},
booktitle = {ICML},
pages = {41–48},
numpages = {8},
location = {Montreal, Quebec, Canada},
}

\newpage
\appendix
\section*{Supplementary Material}

This supplementary document is constructed in the following structures:
\begin{itemize}
    \item In Section A, we detail the used refinement prompt in the third stage of PAG for key phrase generation.
    \item In Section B, we report the human evaluation results of several representative methods.
    \item In Section C, we illustrate the details of hyper-parameters occured in our method.
\end{itemize}

\section*{A. Refinement Prompt} \label{sec:refinement_prompt}
\begin{figure*}[t!]
\centering
\includegraphics[width=1.0\linewidth]{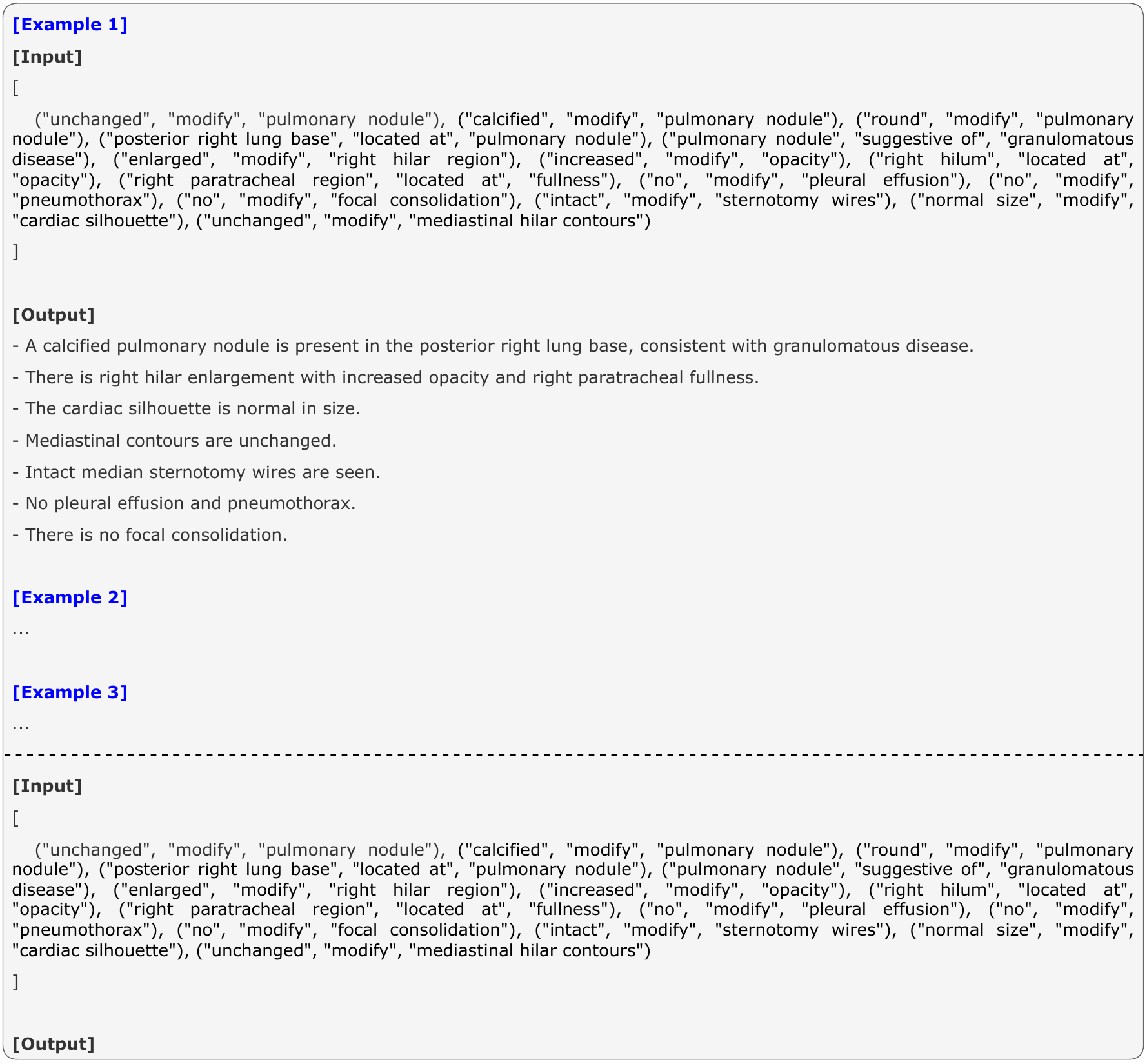}
\vspace{-0.8em}
\caption{\textbf{Illustration of the refinement prompt used in the third stage of PAG.} The system prompt (above the dashed line) provides several few-shot demonstrations to demonstrate the phrase generation process via in-context learning, guiding the Large Language Model (LLM) to convert structured entity-relation tuples into coherent clinical phrases. The user prompt (below the dashed line) then supplies the new set of tuples to be processed.
}

\label{fig:refinement_prompt}
\end{figure*}

In Figure \ref{fig:refinement_prompt}, we show the refinement prompt used in the third stage of PAG to transform the entity-relation triplets into coherent phrases in forms of natural language.
To design such a prompt, our motivation is two-fold.
First, we observe that general-purpose LLMs do not inherently acquire the capability to translate a list of radiological entity-relation tuples into clinically fluent phrases.
While fine-tuning a dedicated model for this specific task is possible, it normally requires significant computational and data annotation costs.
Thus, In-Context Learning (ICL) presents a highly effective and resource-efficient alternative, enabling the LLM to perform this complex translation task without extra parameter updates.
Second, there exists a fundamental structural misalignment between the raw tuples (entities and relations) and the free-text format of complete radiology reports.
The core purpose of this refinement prompt is therefore to transform the structured texts into high-quality descriptive phrases.
These generated phrases are grammatically coherent and semantically aligned with both inputs and outputs of the LLMs in natural language form, serving as precise alignment hints that guide the main model to establish anatomically-grounded correspondences.

Specifically to prepare the in-context examples in this prompt, we construct a curated pool of high-quality examples.
To achieve this, we recruit three board-certified radiologists to manually convert the RadGraph-extracted tuples into a set of gold-standard descriptive phrases.
Once the gold-standard phrases are annotated, we then randomly sample $3$ distinct examples from this annotated pool to serve as the few-shot in-context demonstrations when processing each data sample from the training set.
\begin{table}[h!]
\centering
\setlength{\tabcolsep}{2pt}
\scalebox{0.85}{\begin{tabular}{l l cc}
    \toprule
    \multirow{2}{*}{\textbf{Method}} & \multirow{2}{*}{\textbf{Category}} & \multicolumn{2}{c}{\textbf{Human Evaluation}} \\
    \cmidrule(lr){3-4}
    & & Fluency & F.C. \\
    \midrule
    R2Gen~\cite{chen-etal-2020-generating}         & From-Scratch      & 1.42 & 1.15 \\
    R2GenGPT~\cite{wang2023r2gengpt}     & SFT      & 1.75 & 1.48 \\
    LLM-RG4~\cite{wang-etal-2025-llm-rg4}      & SFT      & 1.82 & 1.65 \\
    \midrule
    \textbf{\textsc{S2D-Align} (Ours)} & \textbf{PAG} & \textbf{1.89} & \textbf{1.78} \\
    \midrule
    \textit{Human Report (Oracle)}  & \textit{N/A}             & \textit{2.00} & \textit{2.00} \\
    \bottomrule
\end{tabular}}
\caption{Human evaluation results on \textsc{MIMIC-CXR} \cite{johnson2019mimic} from perspectives of grammatical fluency and factual correctness \textbf{(F. C.)}.}
\label{tab:human_evaluation}
\end{table}

\begin{table}[h!]
\centering
\resizebox{\columnwidth}{!}{%
\begin{tabular}{l|c}
\toprule
\textbf{Hyper-Parameter} & \textbf{Value} \\
\midrule
\rowcolor{lightgray}
\multicolumn{2}{c}{\textbf{\textit{General Settings}}} \\
\midrule
Optimizer & AdamW \\
AdamW $\beta$s & $(0.9, 0.98)$ \\
Weight Decay & $0.05$ \\
Batch Size & $64$ \\
LR Scheduling & Cosine Annealing w/ Warmup \\
Warmup Steps & $1,000$ \\
\midrule
\rowcolor{lightgray}
\multicolumn{2}{c}{\textbf{\textit{PAG Stage-specific Settings}}} \\
\midrule
Stage 1 (Epochs / LR) & $8$ / $3\times 10^{-4}$ \\
Stage 2 (Epochs / LR) & $5$ / $1\times 10^{-4}$ \\
Stage 3 (Epochs / LR) & $3$ / $5\times 10^{-5}$ \\
\midrule
\rowcolor{lightgray}
\multicolumn{2}{c}{\textbf{\textit{SMA Architecture}}} \\
\midrule
Memory Queries ($N_{mem}$) & $64$ \\
Hidden Dimension ($D_v$) & $768$ \\
\bottomrule
\end{tabular}
}
\caption{Hyper-parameter settings of \textsc{S2D-Align}.}
\label{tab:hyperparams}
\end{table}

\section*{B. Human Evaluation Results} \label{sec:human_evaluation}

To assess the clinical utility of our generated reports beyond automated metrics, we conduct a human evaluation with expert radiologists, with the results on \textsc{MIMIC-CXR} \cite{johnson2019mimic} presented in Table \ref{tab:human_evaluation}.
For this study, generated reports were scored from $0$ (very poor) to $2$ (perfect) on two crucial axes: grammatical fluency and factual correctness (F.C.).
The results clearly show that \textsc{S2D-Align} is significantly preferred by human experts, outperforming all competing methods and achieving scores closest to the human-written oracle reports. 
Notably, while standard SFT methods like LLM-RG4 achieve high fluency ($1.82$), factual correctness of their produced reports is quite limited($1.65$), highlighting the critical limitation of coarse-grained alignment by SFT that only simulates the style of reports but often fails to accurately ground specific clinical findings.
In contrast, the superior F.C. score of \textsc{S2D-Align} ($1.78$) provides compelling evidence that PAG successfully enforces the anatomically-grounded alignment crucial for trustworthy report generation. 
This experiment strongly supports our central hypothesis that explicitly targeting factual correctness through a progressive learning strategy is a pivotal step toward building genuinely reliable LLMs for RRG.

\section*{C. Hyper-Parameter Settings} \label{sec:hyperparameter_settings}

In Table \ref{tab:hyperparams}, we detail the hyper-parameter settings in \textsc{S2D-Align}, so as to provide sufficient details for the purpose of experimental reproduction.

\end{document}